%% file: 0-main.tex
\documentclass[final,3p,times,twocolumn]{elsarticle}

\usepackage{amssymb}
\usepackage{amsmath}

\usepackage{xcolor}
\usepackage{lscape}
\usepackage{multirow}
\usepackage{xurl}

\journal{Journal of Systems and Software}

\begin{document}

\begin{frontmatter}



\title{Architecture for a Trustworthy Quantum Chatbot} 


\author[aff1]{Yaiza Aragonés-Soria\corref{cor1}}
\author[aff1,aff2]{Manuel Oriol}

\affiliation[aff1]{organization={Constructor Institute of Technology},
            addressline={Rheinweg 9}, 
            city={Schaffhausen},
            postcode={8200}, 
            country={Switzerland}}

\affiliation[aff2]{organization={Constructor University},
            addressline={Campus Ring 1}, 
            city={Bremen},
            postcode={28759}, 
            country={Germany}}

\cortext[cor1]{yaiza.aragonessoria@gmail.com}
\begin{abstract}
Large language model (LLM)-based tools such as ChatGPT seem useful for classical programming assignments. 
The more specialized the field, the more likely they lack reliability because of the lack of data to train them.
In the case of quantum computing, the quality of answers of generic chatbots is low.

C4Q is a chatbot focused on quantum programs that addresses this challenge through a software architecture that integrates specialized LLMs to classify requests and specialized question answering modules with a deterministic logical engine to provide trustworthy quantum computing support.
This article describes the latest version (2.0) of C4Q, which delivers several enhancements: ready-to-run Qiskit code for gate definitions and circuit operations, expanded features to solve software engineering tasks such as the travelling salesperson problem and the knapsack problem, and a feedback mechanism for iterative improvement. 

Extensive testing of the backend confirms the system’s reliability, while empirical evaluations show that C4Q~2.0’s classification LLM reaches near-perfect accuracy.
The evaluation of the result consists in a comparative study with three existing chatbots highlighting C4Q~2.0’s maintainability and correctness, reflecting on how software architecture decisions, such as separating deterministic logic from probabilistic text generation impact the quality of the results.

\end{abstract}




\end{frontmatter}



\input{1-introduction}

\input{2-state-of-the-art}

\input{3-features}

\input{4-architecture}

\input{5-results-and-evaluation}

\input{6-conclusions-and-future-work}

\section*{Acknowledgment}
The authors thank Ilgiz Mustafin and Joaquim Soares for their help in the C4Q’s deployment.

\section{Data Availability}
The dataset supporting this study is openly available in \cite{c4q_dataset}. It includes all data and code accompanying the submission, ensuring full transparency and reproducibility of the results.

\bibliographystyle{elsarticle-num-names} 
\bibliography{biblio}

\end{document}

%% file: 1-introduction.tex
\section{Introduction} \label{s:introduction}
Quantum software engineering, which lies at the intersection of quantum physics and computing, is gradually emerging as a distinct field of research. 
This growth is visible as the number of workshops in the field~\cite{palnqc24, QSE2025, esecfse2024Workshops} as well as the gradual integration of quantum computing into software engineering curricula~\cite{masterPrinceton, masterETH, masterUQ}.
Despite this interest, it is still difficult for software engineers to learn how to use quantum computing. 
This poses a significant challenge for learners due to its reliance on advanced mathematics and concepts far removed from everyday experiences. 
One approach to make quantum computing more accessible is to develop more understandable programming languages that simplify quantum concepts for programmers~\cite{DBLP:conf/icse-qse/VargaAO24}.
Another promising solution is a chatbot that can generate and explain code in a quantum programming language.
Yet, a recent analysis of ChatGPT 3.5 highlighted the difficulty of ensuring trustworthy and reliably correct answers in quantum computing~\cite{our_paper}, exposing critical risks when technical accuracy is paramount.
ChatGPT 3.5 frequently provided misleading or blatantly incorrect answers.

In response to these limitations, we developed C4Q\footnote{\url{https://chatbot4quantum.org}} (denoted as C4Q 1.0 for the rest of the text)~\cite{c4q}, a chatbot with a specialized architecture that emphasizes trustworthy and accurate quantum computing answers.
Unlike generic chatbots, which can produce misleading or incorrect responses, C4Q 1.0 integrates a classification Large Language Model (LLM), a Ques\-tion-Ans\-we\-ring (QA) LLM, and a specialized logical engine. 
This software architecture aims to ensure correctness and explainability by isolating the statistical generation of text from the core logic that computes quantum operations. This design is particularly relevant for building trustworthy software, as it mitigates the “black-box effect” often seen in AI-driven solutions, offering a more transparent and auditable approach.

This article extends the foundational work presented on C4Q 1.0~\cite{c4q} by introducing C4Q 2.0, an updated version of C4Q that incorporates additional features and enhancements to broaden its capabilities and reliability. 
Notably, C4Q 2.0 now provides ready-to-run Qiskit code alongside its textual explanations, bridging the gap between conceptual learning and hands-on experimentation $-$ an essential component in building confidence and trust in quantum software. 
Additionally, C4Q 2.0 tackles advanced quantum-based solutions to well-known software engineering problems, such as the Travelling Salesperson Problem (TSP) and the Knapsack Problem (KP). By demonstrating quantum methods for these canonical challenges, C4Q 2.0 provides both software engineers and quantum novices with a more explainable and verifiable pathway into practical quantum computing.
These improvements cater to software engineers and non-experts alike, offering a more comprehensive learning experience by connecting quantum theory with real-world problem-solving. 

Further aligning with the needs of trustworthy and evolving software, C4Q 2.0 includes a feedback mechanism that allows users to rate and comment on its responses. This feature fosters iterative improvement, ensuring that C4Q 2.0’s architecture remains aligned with emerging regulatory standards and user needs for transparency, reliability, and auditability.
To validate C4Q 2.0’s trustworthiness from an architectural perspective, we include a head-to-head comparison of its answers with other chatbots. This empirical evaluation illustrates how C4Q’s specialized design strategies, such as its deterministic logical engine and template-based code generation, outperform more generic solutions in both code correctness and maintainability.
Ultimately, C4Q 2.0 exemplifies how software architecture design can address the demands of accuracy, maintainability, and compliance in quantum computing tools.

This article is divided as follows. Section \ref{s:related_work} presents the state of the art. Section \ref{s:evolution} describes C4Q's features. Section \ref{s:architecture} presents the architecture of C4Q 2.0. Section \ref{s:results_and_evaluation} evaluates the performance of C4Q 2.0. Section \ref{s:conclusions} concludes and presents future work.

%% file: 2-state-of-the-art.tex
\section{State of the Art} \label{s:related_work}
Ensuring a trustworthy software architecture has become a priority for AI-driven systems, emphasizing the importance of transparency, auditability, and compliance in software systems~\cite{LANGER2021103473, 10.1145/3561048, 10.1145/3308532.3329441, EvalPlus}. 
This challenge is particularly crucial in programming-related tasks, where LLMs are largely used for code generation~\cite{review_LLMs_code_generation, 8823632, 10.1145/3183519.3183540} such as
the insertion of automatically proven contracts~\cite{DBLP:journals/tse/CarrLP17}, or the guidance on APIs~\cite{8115628}.

In quantum computing, precision and accuracy is paramount: errors in gate definitions or circuit setups can have large downstream effects. Despite these demands, industry has also explored AI-driven tools that focus on simplifying the learning and development process. For instance, Microsoft's Copilot in Azure Quantum~\cite{chatbot_azure} is a ``quantum-focused chatbot designed to assist in code writing and enhance understanding of quantum concepts'' by utilizing Q\#.
In contrast to C4Q’s design, Copilot leverages GPT-4’s probabilistic generation to clarify complex quantum phenomena. While this approach lowers barriers to a challenging field, the probabilistic nature of generative models can complicate trustworthiness by introducing a gap between user expectations and actual behaviour. Moreover, generic LLMs have demonstrated inconsistent performance in quantum-related queries~\cite{our_paper, QiskitHumanEval}, raising concerns about trusting black-box architectures for specialized domains such as quantum software engineering.

In the educational context, many see in generative AI a transforming technology for teaching and research~\cite{murugesan2023, damian2023, yilmaz2023, lim2023}, leading to works such as Kotovich \textit{et al.}~\cite{fisee23Kotovich}, who evaluated how well ChatGPT can generate fundamental algorithms.
Nonetheless, for quantum-specific tasks, ChatGPT-3.5 and ChatGPT-4 do not consistently maintain the accuracy required for reliable learning experiences~\cite{chatgpt4_study}.
Despite these evaluations, some attempts have been made to use packed GTP4 to "demystify quantum computing".\footnote{\url{https://www.yeschat.ai/gpts-2OTogwe6nW-Quantum-AI-Chatbot}}$^{,\thinspace}$\footnote{\url{https://sodoherty.ai/2021/03/}}
The tools however do not live up to the expectations and lack the accuracy that one needs to learn safely.

As a notably early attempt to provide an accurate tool, C4Q~\cite{c4q} provided a solution by separating probabilistic text generation from core quantum operations through a classification LLM, a QA LLM for parameter extraction, and a specialized logical engine for deterministic computations. 
This architecture embodies trustworthy software principles such as modularity, traceability, and fine-grained auditing~\cite{10.5555/2392670}.
Similar specialized LLM solutions exist for non-quantum domains, such as the plugin of Wolframalpha for ChatGPT~\cite{chatGPT}, which promises ``powerful computation, accurate math, curated knowledge, real-time data and visualization'', or CrunchGPT~\cite{Kumar_2023} for scientific machine learning.

Building on the modular nature of C4Q, we refer to the first release~\cite{c4q} as C4Q 1.0, and introduce an updated version, C4Q 2.0, which offers new features, additional quantum capabilities and reinforced reliability.

%% file: 3-features.tex
\section{C4Q’s Features} \label{s:evolution}
This section provides an overview of C4Q's core functionalities, emphasizing how its architecture supports trustworthy and maintainable quantum software. 
First, we summarize the foundation features from C4Q~1.0~\cite{c4q}, highlighting enhancements that improve both transparency and practical usability through ready-to-run code integration. 
Next, we describe the newly introduced capability for solving classical programming using quantum methods, specifically addressing the TSP and KP.
Finally, we outline the integrated user feedback mechanism designed to facilitate continuous improvement, reinforcing the user-centered and auditable nature of C4Q's architectural design.

\subsection{Foundational Features} \label{s:enhanced_responses}
C4Q was initially developed as a targeted solution to the reliability issues identified in generic chatbots, such as ChatGPT~\cite{our_paper} for quantum computing. 
In its original version (C4Q 1.0), the chatbot focused on delivering accurate responses to queries involving a predefined set of fundamental quantum gates, including the identity, the Pauli gates, the $S$ and $S^\dag$ gates, the Hadamard gate, phase gates, rotations, the CNOT gate, CZ gate, and the SWAP gate.

For each quantum gate within this set, C4Q 1.0 handled three distinct types of queries reliably:
\begin{enumerate}
    \item \emph{Defining} the quantum gate, providing a comprehensive description of the specific quantum gate.
    \item \emph{Drawing} the quantum gate, by generating a circuit representation of the quantum gate.
    \item \emph{Applying} the gate on an initial state given by the user, calculating and presenting the resulting state.
\end{enumerate}

To further enhance accuracy, C4Q~1.0 includes a verification step: before generating an answer, C4Q requests confirmation to avoid misunderstandings and continually enhance the training of its underlying models. This design exemplifies a trustworthy software architecture by fostering transparency, traceability, and user-driven improvements.

C4Q 2.0 extends the foundational tasks of C4Q~1.0 by adding practical hands-on capabilities that further support the trustworthiness and explainability of C4Q. 
While gate definitions remain textual to clearly convey fundamental concepts, the tasks involving circuit representation and gate applications are now supplemented with ready-to-run Qiskit code including explanatory comments. 
This code is generated from internal templates and
can be easily copied with a single button click, enabling seamless use on the user’s computer. 
The inclusion of ready-to-run code ensures not only correctness and transparency but also maintainability, as templates for code snippets facilitate easier adaption to evolving quantum software libraries like Qiskit.

\begin{figure}[htbp]
\centerline{\includegraphics[scale=0.35]{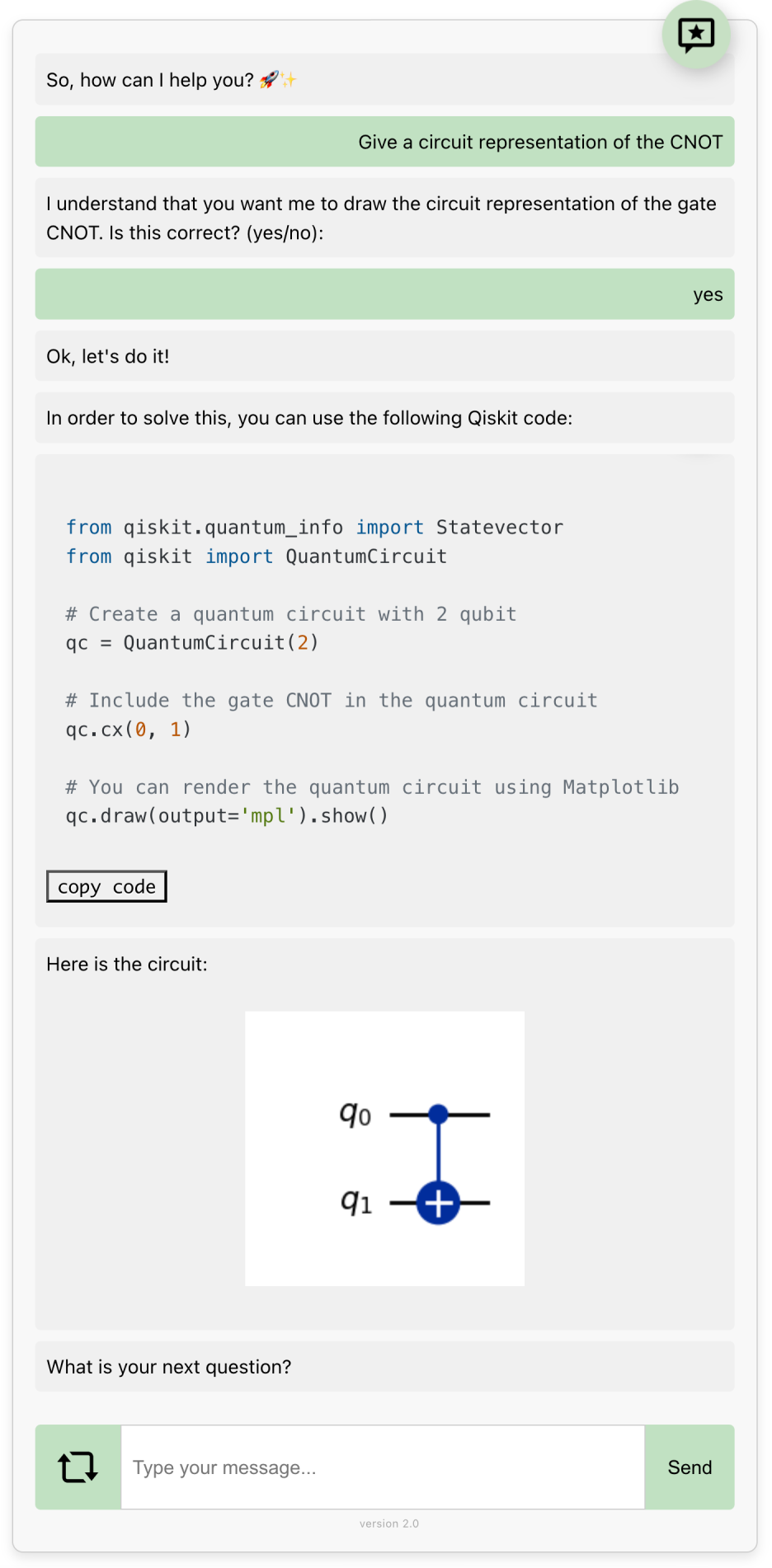}}
\caption{Feature enhancement in C4Q 2.0 showcasing a dual response to a user’s request for a CNOT gate representation: a circuit diagram, and the corresponding Qiskit code.}
\label{f:draw_with_code}
\end{figure}

Figure~\ref{f:draw_with_code} illustrates how C4Q 2.0 addresses a user request for a CNOT gate’s circuit representation. 
In response, the chatbot provides both the visual diagram and the annotated Qiskit code necessary to reproduce it, giving users a comprehensive and verifiable answer. 

The new dual response of C4Q bridges theoretical insight and hands-on code, empowering users to transition smoothly from understanding quantum gates conceptually to implementing them practically. 
In this way, C4Q~2.0 not only reinforces the educational aspects of quantum computing but also exemplifies an architecture designed explicitly to foster trust, transparency, and maintainability, crucial for reliable software in the complex domain of quantum computing.

\subsection{Solving Programming Problems} \label{s:solving_SE}

Recognizing the potential of quantum computing to address optimization tasks with significant complexity, we incorporated fully coded quantum algorithms as part of C4Q 2.0’s trustworthy solution offerings. This section introduces the C4Q’s new capability to solve the Travelling Salesperson  Problem (TSP) and the Knapsack Problem (KP), reflecting the expandable nature of C4Q’s architecture for addressing real-world programming requests.

\subsubsection{Travelling Salesperson Problem (TSP)}\label{s:tsp}
The objective of the TSP is to determine the most efficient route for a salesperson who must visit $N$ different cities exactly once before returning to the starting point. 
The sum of distances between cities serve as the cost function to be minimized. 
The TSP is classified as an NP-hard problem in combinatorial optimization~\cite{Korte2008}, meaning that the time required to find an exact solution increases exponentially with $N$.

The general TSP and its various adaptations find applications across multiple domains, including DNA sequencing, logistics and transportation, manufacturing and production planning, and network design and optimization~\cite{Punnen2007}. 
Although several classical methods exist for solving the TSP $-$ such as brute-force algorithms~\cite{applegate2011traveling}, and branch and bound techniques~\cite{42fd065c-22f8-35d7-a4cc-51ed7f70cf3f} $-$ the time required to obtain an exact solution grows exponentially with the number of cities. 
Heuristic methods~\cite{RePEc:inm:ormnsc:v:10:y:1964:i:2:p:225-248} can accommodate larger instances of the problem, but they do not guarantee optimal solutions, and verifying their optimality can be challenging due to the NP-hard nature of the problem.

Research in quantum algorithms for solving the TSP remains active, with promising proposals aimed at achieving practical advantages in both speed and accuracy compared to classical algorithms. 
For instance, Srinivasan \textit{et al.}~\cite{srinivasan2018efficientquantumalgorithmsolving} suggest a quadratic speedup over classical brute-force methods using phase estimation techniques, while Bang \textit{et al.}~\cite{Bang_2012} present a heuristic algorithm that generalizes Grover's search to address the TSP. 
Additionally, adiabatic quantum computation has been proposed as a potential framework for solving the TSP~\cite{Kieu_2019}.

Given the TSP’s relevance and its potential as an educational tool to demonstrate the capabilities of quantum computing, we have integrated it into C4Q 2.0’s feature set. 
C4Q 2.0 accepts user-defined TSP instances, generates ready-to-run Qiskit code to solve the problem, and provides the solution of the TSP. 
C4Q 2.0 converts the TSP into a quadratic problem and employs a Variational Quantum Eigensolver (VQE)~\cite{VQE} to find the optimal route.

\begin{figure*}[htbp]
\centerline{\includegraphics[scale=0.30]{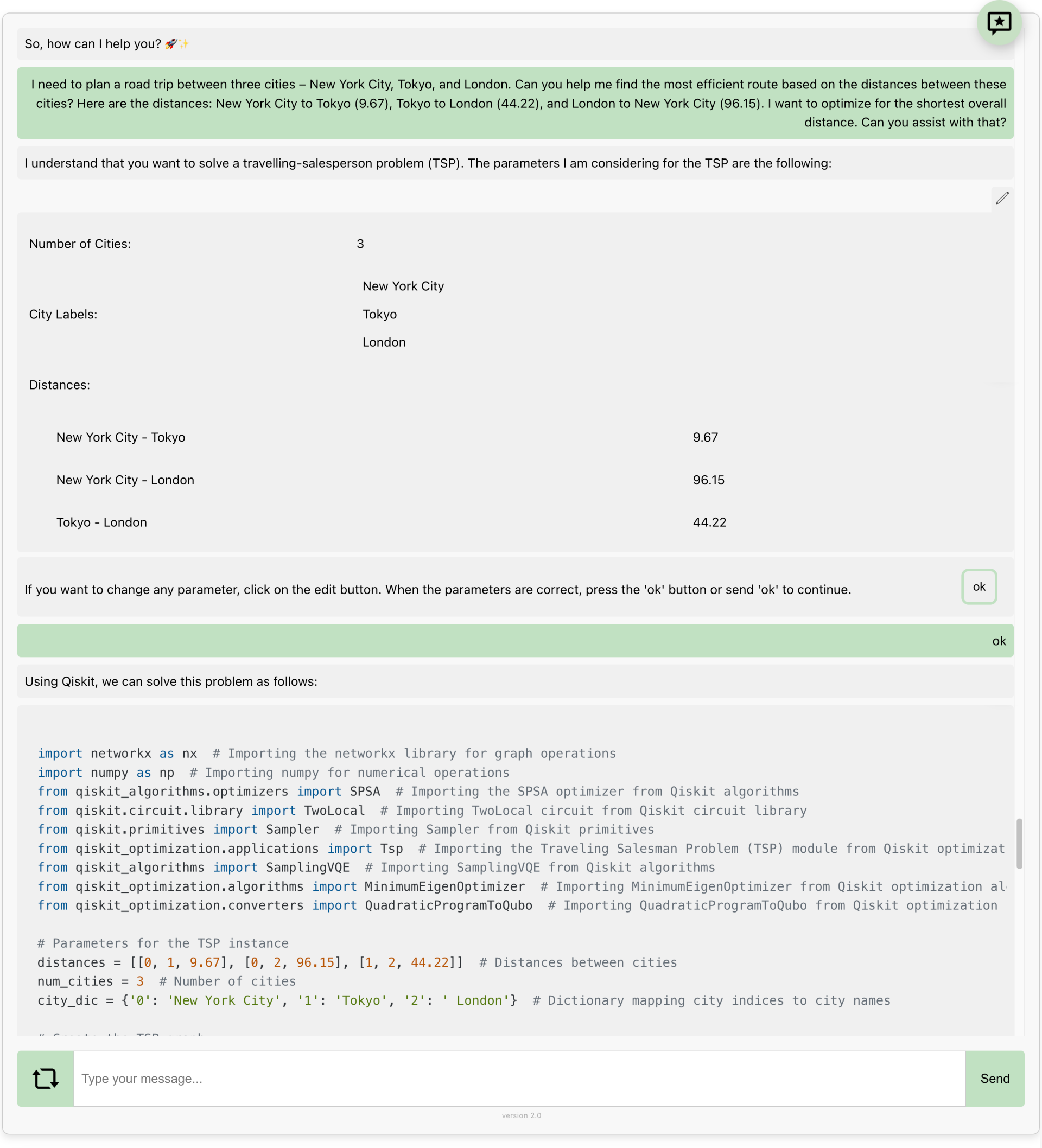}}
\caption{C4Q 2.0 user interaction for solving a TSP. C4Q 2.0 extracts parameters from the user’s question and generates Qiskit code based on the confirmed parameters.}
\label{f:tsp}
\end{figure*}

Figure~\ref{f:tsp} illustrates an example of user interaction in which C4Q 2.0 solves a TSP. Upon receiving the user’s query, C4Q 2.0 analyses the input, extracts the relevant parameters, and presents them to the user for confirmation or modification. Once the parameters are confirmed, C4Q 2.0 generates ready-to-run Qiskit code to solve the problem and provides an option for the user to initiate the computation in C4Q 2.0's backend. 
By splitting code generation from solution computation, the system aligns with trustworthy design principles, giving users control over runtime decisions and ensuring reproducibility and transparency.
The generated code is easily accessible, with a copy button for quick transfer, allowing seamless execution on the user’s computer if preferred. 
This approach ensures that users who do not require C4Q to give the concrete outcome of the code do not have to wait for lengthy computations.

The incorporation of TSP into C4Q 2.0's capabilities significantly enhances its utility, offering users Qiskit code and computation options. 

\subsubsection{Knapsack Problem (KP)} \label{s:knapsack}
The KP is a classic combinatorial optimization challenge where a set of items, each with an assigned weight and value, must be selected to maximize total value without exceeding a specified weight limit for the knapsack. This problem and its variations have significant real-world applications in fields like finance, resource allocation, and manufacturing, where optimal resource distribution is essential~\cite{knapsack-applications}.

As an NP-hard problem~\cite{knapsack-NP}, KP is of particular interest in quantum computing, where researchers aim to leverage quantum algorithms for potential speed-ups over classical methods.
Reformulating KP as a quadratic unconstrained binary optimization (QUBO) problem enables it to be solved on both quantum annealing~\cite{QA} and gate-based quantum algorithms~\cite{gate-based-algorithms}, such as the Quantum Approximate Optimization Algorithm (QAOA)~\cite{QAOA} and VQE. 
Awasthi \textit{et al.}~\cite{Awasthi_2023} compare the effectiveness of various gate-based quantum algorithms, including QAOA and VQE, with quantum annealing for solving KP. 
Other research has led to specifically tailored quantum algorithms for KPs, highlighting the problem’s relevance in quantum optimization~\cite{christiansen2024quantumtreegeneratorimproves}.

Given the interest in KP within quantum computing and its applicability in software engineering, we have incorporated KP-solving functionality into C4Q 2.0. Similar to the approach taken for the TSP in Section \ref{s:tsp}, when a user inputs a question asking to solve a KP, C4Q 2.0 extracts the parameters of the problem and displays them to the user for verification and modification if needed.
Once confirmed, C4Q 2.0 generates ready-to-run Qiskit code to solve the problem.
In line with trustworthy design principles, C4Q 2.0 offers the option to perform the computation on its backend or locally, enhancing user control and traceability. 
If the user opts to compute the solution in C4Q 2.0's backend, C4Q 2.0 utilizes a QAOA-based quantum optimizer to transform the KP into QUBO format and solve it.

The addition of KP capability not only enriches C4Q’s educational scope but also demonstrates both the extensibility of the C4Q architecture, and the potential of quantum methods to tackle optimization challenges relevant to software engineers.

\subsection{User Feedback Mechanism} \label{s:feedback}
To support continuous improvement and user-driven evolution, C4Q 2.0 includes a feedback mechanism for users to provide ratings (1 to 5 stars) and optionally submit written comments.
The feedback button is visible at the top right corner of the interface, as shown in Figures~\ref{f:draw_with_code} and~\ref{f:tsp}.

By gathering direct input from a diverse user base, this feedback mechanism bolsters transparency and accountability, key pillars of trustworthy software. It also plays a pivotal role in future user studies, guiding refinements and ensuring that C4Q evolves in alignment with real-world needs, best practices, and emerging standards in software engineering.

%% file: 4-architecture.tex
\section{Architecture of C4Q 2.0} \label{s:architecture}
This section provides an integrated view of the C4Q architecture, focusing on both C4Q 1.0 (as introduced in~\cite{c4q}) and C4Q 2.0. We highlight how architectural decisions support trustworthy and explainable software, in line with the requirements for modern systems that must demonstrate transparency, compliance, and reliability.

\subsection{Overview of C4Q 1.0}

\begin{figure*}[htbp]
  \centering
  \includegraphics[width=0.8\linewidth]{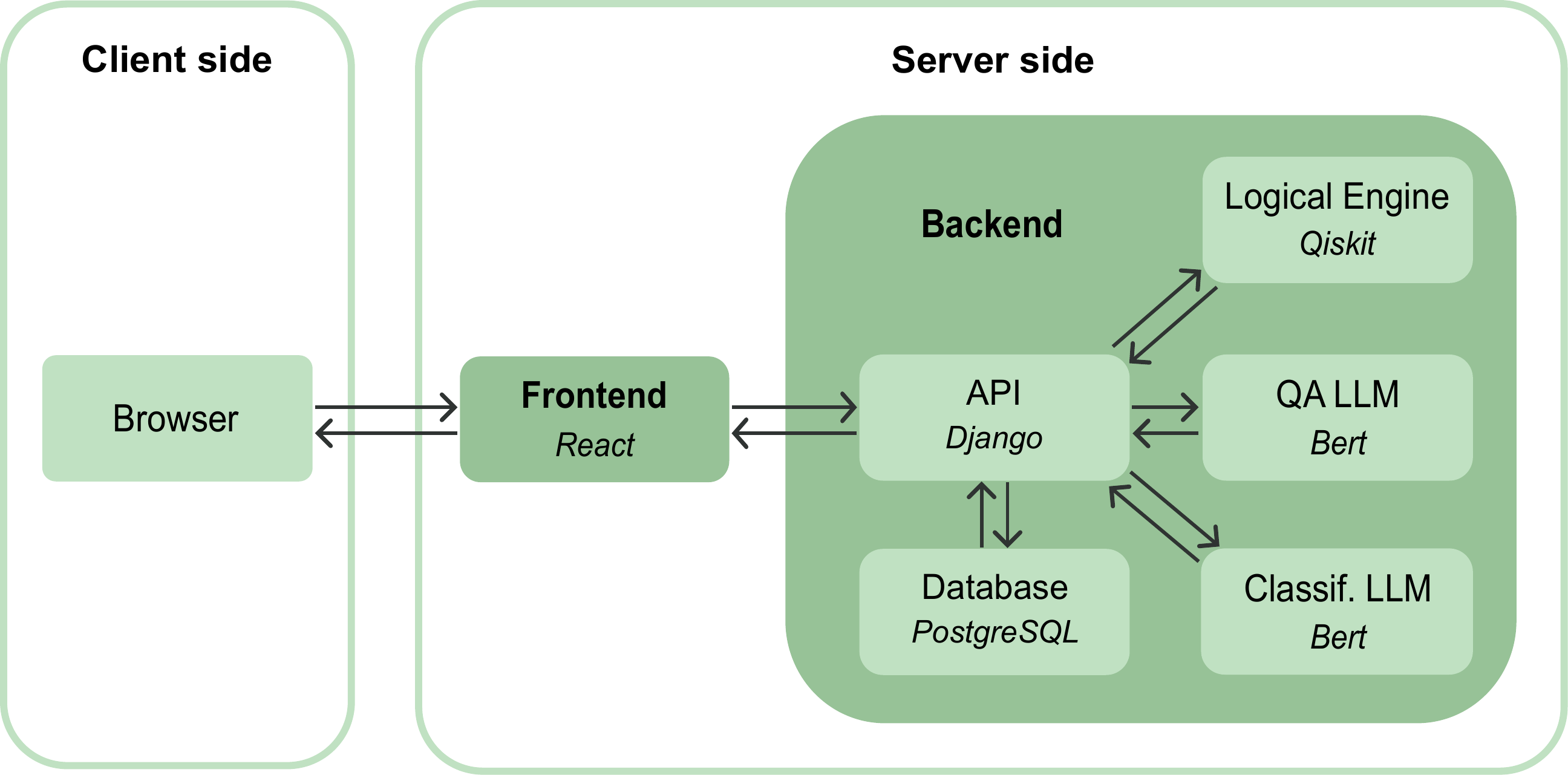}
  \caption{High-level architecture diagram of C4Q.}
  \label{f:architecture_diagram}
\end{figure*}

C4Q 1.0’s architecture is composed of multiple modules that handle different aspects of user requests, as depicted in Figure~\ref{f:architecture_diagram}. 
This modular architecture ensures each component handles a focused set of responsibilities, thereby fostering traceability and maintainability.

The frontend, developed using React, provides a straightforward and user-friendly interface. In the backend, five modules contribute to C4Q 1.0’s functionality:
\begin{enumerate}
    \item \textbf{API.} This module acts as an intermediary between the frontend and backend components. Built with Django,\footnote{\url{https://www.djangoproject.com}} it manages communication tasks such as routing user requests, retrieving data, and interacting with the classification and QA LLMs. Its design enables future auditing and extensibility, both critical for building trustworthy systems that may require regulatory compliance.

    \item \textbf{Database.} The database leverages PostgreSQL with the psycopg2 adapter to store user and message information. By default, C4Q 1.0 deletes full conversations at session end to maintain a streamlined database while retaining essential user questions for continuous classification improvements. This policy balances data privacy and model enhancement needs.

    \item \textbf{Classification LLM.} A fine-tuned BERT model classifies user questions into three tasks: defining a quantum gate, drawing a quantum gate, or applying a quantum gate. By focusing on early-task segregation, C4Q 1.0 reduces reliance on purely probabilistic outputs, thereby improving system reliability.

    \item \textbf{QA LLM.} Another fine-tuned BERT-based Question ans Answer (QA) model extracts critical parameters, such as the phase shift for phase gates or the rotation angle and axis for rotation gates. By limiting the scope to parameter extraction rather than full answer generation, this module increases explainability and accuracy.

    \item \textbf{Logical Engine}. After classification and parameter extraction, the logical engine handles core quantum computations. It integrates with Qiskit to produce deterministic, mathematically correct answers rather than relying on LLM-generated outputs. This design further strengthens trust by confining LLM usage to classification and parameter extraction, while quantum computations remain deterministic. 
    
\end{enumerate}

Therefore, C4Q 1.0 presents a modular design that promotes accurate quantum computations and targeted use of LLMs. The next subsection describes how C4Q 2.0 expands on this architecture to deliver new functionalities while upholding principles of trust and transparency.

\subsection{Architectural Enhancements in C4Q 2.0}
Section \ref{s:evolution} describes the new capabilities of C4Q 2.0 in detail, including outputting Qiskit code, solving TSPs and KPs, and the newly added feedback feature. This section focuses on the architectural updates that the new capabilities required.

Building on C4Q 1.0’s modular foundation, C4Q 2.0 introduces several architectural and feature enhancements to broaden capabilities while preserving user trust and system transparency. Specifically, these updates revolve around:
\begin{itemize}
    \item \textbf{Extended Classification LLM.} The classification LLM now recognizes five categories: defining a quantum gate, drawing a quantum gate, applying a quantum gate, and solving the Travelling Salesperson Problem (TSP) or the Knapsack Problem (KP). This change maintains the original pipeline’s reliability while covering additional user requests.

    \item \textbf{Enhanced QA LLM.} The QA model is fine-tuned to handle extra parameters for TSP (for example, city names and distances) and KP (for example, item weights and values). By retaining a focus on structured parameter extraction rather than probabilistic text generation, C4Q 2.0 ensures traceable logic for these new problems.

    \item \textbf{Templated Qiskit Code Generation.} In contrast to the more error-prone approach of having an LLM generate code from scratch, C4Q 2.0 employs a template-based mechanism. The QA LLM populates placeholders in curated Qiskit code snippets with the extracted parameters, reducing syntax and logic errors, reinforcing the system’s trustworthiness, and making maintenance more straightforward. For instance, if Qiskit updates to a new version, developers can revise the template rather than retraining an LLM with extensive new examples.

    \item \textbf{Additional Problem-Solving Support.} Beyond quantum gates, the logical engine can now convert user TSP or KP data into a quantum optimization model (for example, QUBO) and apply algorithms such as the Variational Quantum Eigensolver (VQE) or the Quantum Approximate Optimization Algorithm (QAOA). The API routes these specialized queries to routines that manage parameter extraction, template filling, and quantum computation.

    \item \textbf{Feedback Mechanism for Continuous Improvement.} C4Q 2.0’s Django-based feedback model collects user ratings and written comments for later analysis. This process supports user-driven refinements, addressing socio-technical aspects of trust by incorporating stakeholder feedback into the platform’s evolution.
\end{itemize}

By covering an expanded range of tasks, integrating templated code generation, and implementing direct feedback loops, C4Q 2.0 preserves the deterministic quantum computations introduced in its predecessor while delivering broader functionality. The next subsection explains how these choices align with requirements for transparency, auditing, and maintainability in trustworthy software.

\subsection{Trustworthiness and Compliance Considerations}
Given the increasing legislative and industry push toward trustworthy software, C4Q 2.0’s architecture incorporates specific transparency and explainability mechanisms:

\begin{itemize}
    \item \textbf{Separation of Concerns.} Classification, parameter extraction, and logical computation are handled by distinct modules, reducing black-box risks and enabling more granular audits.

    \item \textbf{Auditable Interactions.} Essential interaction data is retained in the database, supporting audits, reproducibility, and potential compliance with AI transparency regulations.
    
    \item \textbf{Explainability.} By returning both textual explanations and Qiskit code, the system offers users a practical way to inspect and verify each step, thereby promoting accountability and stakeholder confidence.
    
    \item \textbf{Feedback-Driven Updates.} The integrated feedback mechanism ensures the architecture can evolve based on real-world needs, user satisfaction, and emerging best practices in software engineering.
\end{itemize}

These measures collectively help C4Q 2.0 comply with modern guidelines for trustworthy software by ensuring transparency, enabling effective audits, and providing clear explanations of how computations are derived. The architecture thus remains flexible enough to accommodate evolving compliance requirements while sustaining user trust.

%% file: 5-results-and-evaluation.tex
\section{Results and Evaluation} \label{s:results_and_evaluation}
This section examines how C4Q~2.0's architecture upholds trustworthiness and transparency through both internal testing and comparative analysis. 
We begin by assessing the stability of its backend components and the performance of its fine-tuned LLMs. 
Next, we analyse the system's maintainability, demonstrating how its modular design enables efficient updates and extensions with minimal computational overhead. 
Finally, we compare C4Q~2.0 against other chatbots, highlighting its architectural benefits related to accuracy, regulatory compliance, and long-term sustainability.

\subsection{Backend and LLM Performance}\label{ss:backend_llm_performance}

To ensure that C4Q~2.0 meets the reliability demands of trustworthy software, we performed 189 unit tests using the \texttt{pytest} framework, ensuring that each function and method within the backend was individually tested. 
The backend testing report~\cite{c4q_dataset} confirms that all tests passed successfully, demonstrating the stability and reliability of the backend components. This result is in line with the system’s emphasis on maintainability and trustworthiness, as a well-tested backend forms a solid foundation for compliance with audit and certification processes.

\begin{figure}[htbp]
\centerline{\includegraphics[scale=0.75]{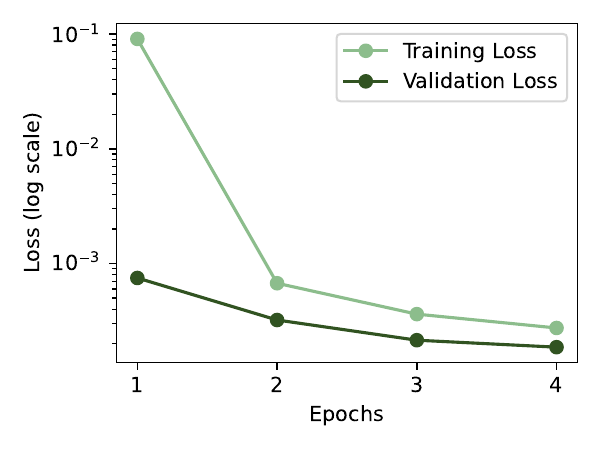}}
\caption{Training and validation loss of the classification LLM.}
\label{f:class_metrics_loss}
\end{figure}

For evaluating the classification LLM, we computed both the training and evaluation loss, as well as the training and validation accuracy. 
Figure~\ref{f:class_metrics_loss} shows the loss progression across epochs. The training and evaluation loss dropped to below $10^{-3}$ by the second epoch achieving a value equal to $2.74\cdot10^{-4}$ and $1.87\cdot10^{-4}$, respectively, by the fourth epoch. 
The classification model achieved perfect accuracy, with both training and validation accuracy consistently reaching 1.00 from the first epoch onward. 
Such robust performance in distinguishing question types is pivotal for supporting trustworthy and transparent interactions at the architectural level, since an accurate classification module helps direct user queries to the correct logic path.

\begin{figure}[htbp]
\centerline{\includegraphics[scale=0.75]{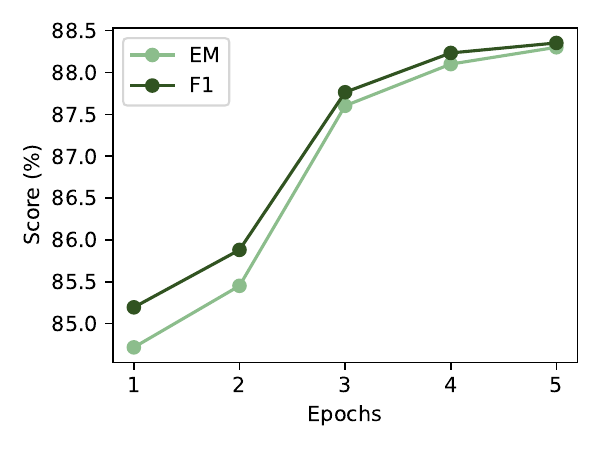}}
\caption{Exact match, EM, and F1 (in percentage points) of the QA LLM.}
\label{f:qa_metrics}
\end{figure}

To assess the QA LLM’s accuracy in extracting relevant information, we used both exact match, EM, and F1\footnote{Throughout this article, the F1 is presented in percentage points.} metrics, which are presented in Figure \ref{f:qa_metrics}. 
The results show that both the EM and F1 reach over 87.5\% by the third epoch and gradually increase with additional epochs, plateauing at around 88.3\%. 
While gains from further fine tuning appear limited, the QA LLM’s performance remains sufficient for most quantum queries, reinforcing C4Q~2.0’s goal of providing verifiable and transparent information.

To better understand where the QA LLM makes incorrect predictions, we analyzed which types of questions are most prone to errors. The QA LLM has been fine-tuned to answer the following nine questions:
\begin{itemize}
    \item What is the phase shift?
    \item What is the angle of the rotation?
    \item What is the axis of the rotation?
    \item Which cities does the person want to visit?
    \item What is the distance between \texttt{city1} and \texttt{city2}?
    \item Which items can be selected?
    \item What is the maximal weight?
    \item What is the weight of \texttt{item}?
    \item What is the value of \texttt{item}?
\end{itemize}

\begin{figure}[htbp]
\centerline{\includegraphics[scale=0.75]{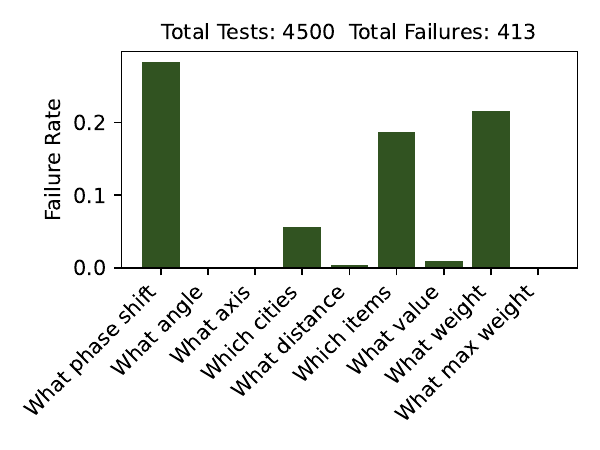}}
\caption{Failure rate for each question type in the QA LLM, calculated as the ratio of incorrect answers of each question to the total number of times each question was asked.}
\label{f:qa_metrics_failures_rates}
\end{figure}

Each question was asked to the QA LLM at least 292 times, and we recorded the number of times it provided incorrect predictions, i.e., predictions that differ with at least a character (excluding punctuation) from the expected answer. 
Figure \ref{f:qa_metrics_failures_rates} shows the failure rate for each question relative to the number of times this question was asked. Our analysis reveals that the question \emph{What is the phase shift?} yields incorrect answers in almost 30\% of cases. For the questions \emph{What is the weight of \texttt{item}?} and \emph{Which items can be selected?}, the QA LLM makes errors in approximately 20\% of cases. In contrast, questions such as \emph{What is the angle of the rotation?}, \emph{What is the axis of the rotation?}, and \emph{What is the maximal weight?} consistently yield accurate answers, with error rates below 10\% and some even reaching perfect accuracy.

To refine the QA LLM’s predictions, we examined whether the incorrect answers exhibit identifiable patterns. For \emph{What is the phase shift?}, the QA LLM often confuses the phase shift with the initial state, producing answers related to it. However, there are also cases where, despite the presence of initial state information in the context, the QA LLM correctly identifies the phase shift. For the question \emph{What is the weight of \texttt{item}?}, the QA LLM’s errors frequently involve selecting another numeric value present in the context, not corresponding to the item’s weight. In the case of the question \emph{Which items can be selected?}, incorrect answers typically provide only a subset of the relevant items in the KP.

The above observations provide valuable insights into the QA LLM’s performance across various question types and guide our approach for future improvements. Despite these areas for improvement, C4Q~2.0 provides user-centric mitigations: for foundational questions on quantum gates, it confirms its interpretation with the user, and for software engineering problems like the TSP or KP, it allows manual review and correction of parameters. Both features preserve transparency and user control, aligning with best practices for building trustworthy software.

In summary, these evaluations confirm that C4Q~2.0’s modular design yields high backend reliability, stable classification results, and robust (though not flawless) QA performance. By maintaining an architecture that separates classification, question answering, and deterministic computations, C4Q~2.0 enhances explainability and verifiability, whcih are key tenets for compliance and stakeholder trust. The combined results indicate that C4Q~2.0 is well-positioned as a learning and development tool for quantum software engineering, while also offering a solid platform for further innovation and empirical refinement.


\subsection{Maintainability of C4Q's Architecture} \label{s:maintainability}
A key advantatge of C4Q's modular design is its maintainability, which ensures that the system can be efficiently extended to support new types of questions. 

Adding a new query type to C4Q follows a structured and reproducible process, comprising the following steps:
\begin{enumerate}
    \item \textbf{Generating training data for the classification LLM.} This involves writing multiple variations of how a user might phrase the new question.
    
    \item \textbf{Generating training data for the QA LLM.} This requires constructing sets of (context, question, answer) triplets, where:
        \begin{itemize}
            \item The \emph{context} represents a possible user prompt.
            \item The \emph{question} extracts a specific parameter from the prompt.
            \item The \emph{answer} provides the corresponding parameter value.
        \end{itemize}

    \item \textbf{Fine-tuning the LLMs}: Both the classification and QA LLMs must be fine-tuned on the newly generated data.

    \item \textbf{Defining the response logic}: This involves designing the textual response and, when applicable, creating the corresponding Qiskit template to generate code-based answers.

    \item \textbf{Updating the API}: The backend must be informed of the newly added question type.

    \item \textbf{Testing}: Comprehensive testing is conducted to ensure correctness and robustness.
\end{enumerate}

This well-structured process allows for the seamless integration of new functionalities into C4Q. 
The main effort lies in curating high-quality training data, while the actual fine-tuning of the LLMs is computationally feasible. 
For instance, adding the TSP and KP to C4Q required approximately nine hours of fine-tuning per model on a MacBook Pro with an Apple M2 Pro chip and 32 GB of RAM.

Compared to end-to-end LLM-based chatbots, C4Q's modular and hybrid approach significantly improves maintainability. 
The structured addition of new question types ensures that updates are systematic, reducing the need for costly retraining of large-scale models.
This results in a faster, more scalable, and more sustainable solution for a trustworthy quantum chatbot.


\subsection{Empirical Comparison} \label{ss:empirical_comparison}

To further substantiate C4Q’s trustworthiness from an architectural perspective, we compared its answers to those of three other chatbots: ChatGPT with \texttt{openai-o1}, Olama with \texttt{deepseek-coder:33b}, and Deepseek’s chatbot with \texttt{deepseek-r1}. We focused on chatbots or command-line interfaces (CLIs) that can generate Qiskit code and user-facing explanations, mirroring C4Q’s functionality.

We selected \texttt{openai-o1}, \texttt{deepseek-coder:33b}, and \texttt{deepseek-r1} owing to their documented performance on quantum code generation tasks and their availability as either complete chatbots or CLIs. Specifically, \texttt{deepseek-coder:33b} was the second-best performer in~\cite{QiskitHumanEval}, where LLMs were evaluated using the newly introduced Qiskit HumanEval dataset. 
The best overall LLM reported in~\cite{QiskitHumanEval} is not accessible via a chatbot or CLI, making \texttt{deepseek-coder:33b} our top choice. 
In the comparison, we also included \texttt{openai-o1}, a reasoning variant of \texttt{GPT-4o}\cite{openai_web}, due to its recognized effectiveness in generating code\cite{EvalPlus}. Finally, \texttt{deepseek-r1} claims superior performance to \texttt{GPT-4o}\cite{deepseek_web} and, in particular version r1, specialises in reasoning tasks.

In designing our evaluation, we found the Qiskit HumanEval dataset unsuitable for C4Q because it presents prompts made of partial code snippets that must be completed by the LLM. 
Since C4Q’s architecture is not trained to handle incomplete snippets, we created our own set of questions \cite{c4q_dataset}. 
Each question reflects a realistic user query, and we scored the chatbots’ answers according to the methodology proposed in~\cite{our_paper}. Specifically, each answer is labeled as:
\begin{enumerate}
    \item \textit{correct}, if all information (including code) is accurate and complete;
    \item \textit{incomplete}, if the answer is correct but omits required details; or
    \item \textit{incorrect}, if any part of the answer is factually wrong or the code fails to run.
\end{enumerate}
To minimize ambiguity in judging completeness, we established specific requirements per task type (e.g., gate definition must include a matrix representation or basis transformation explanation).

Once the evaluation criteria had been established, we posed the same question to each chatbot once and assessed their responses according to the defined criteria. 
Because the release of Qiskit 1.0.0 introduced a major reorganization of its package structure \cite{Qiskit1.0.0}, we decided to test whether the generated code would run in two distinct scenarios: one using Qiskit 0.46.3 (i.e., a version below 1.0.0), and another using Qiskit 1.3.1 (i.e., a version at or above 1.0.0). Preliminary evidence suggested that the LLMs’ performance varied significantly between these two setups, thus motivating our decision to perform separate evaluations under each version.

\begin{table*}[t]
    \centering
    \begin{tabular}{lcccccc}
        \hline
         & \multicolumn{2}{c}{\% Correct} & \multicolumn{2}{c}{\% Incomplete} & \multicolumn{2}{c}{\% Incorrect}\\
        \cline{2-7}
         Qiskit version & <1.0.0 & >1.0.0 & <1.0.0 & >1.0.0 & <1.0.0 & >1.0.0 \\
        \hline
        C4Q                & 97.92 & 97.92 &  2.08 &  2.08 &  0.00 &  0.00 \\
        \texttt{chatgpt-o1 }        & 61.46 & 25.00 & 14.58 &  2.08 & 23.96 & 72.92 \\
        \texttt{deepseek-r1}        & 59.38 & 13.54 & 14.58 &  8.33 & 26.04 & 78.13 \\
        \texttt{deepseek-coder:33b} & 11.46 &  3.13 & 17.71 &  2.08 & 70.83 & 94.79 \\
        \hline
    \end{tabular}
    \caption{Percentage of correct, incomplete and incorrect answers across Qiskit versions, where we used Qiskit 0.46.3 and Qiskit 1.3.1.}
    \label{t:results}
\end{table*}

Table~\ref{t:results} summarizes the outcomes of our empirical evaluation, revealing that C4Q obtains the best results, with near-perfect accuracy and consistently correct, complete answers across all question types. When testing code snippets using Qiskit $< 1.0.0$, \texttt{chatgpt-o1} and \texttt{deepseek-r1} obtained correctes rates of 61.46\% and 59.38\%, respectively, while \texttt{deepseek-coder:33b} scored remarkably lower. However, for Qiskit $\geq1.0.0$, the accuracy of all chatbots except C4Q dropped to 25\% or lower. This significant decline can be attributed to the fact that most publicly available code, and thus the training data for generic LLMs, targets Qiskit releases preceding version~1.0.0. C4Q’s template-based code generation, however, adapts easily to new Qiskit versions with minimal changes, highlighting the maintainability benefits of its architecture. By contrast, to update the code generated by generic LLMs would require substantial data collection with updated Qiskit code as well as retraining.

Notably, \texttt{deepseek-coder:33b} achieved only 11.46\% correct answers with Qiskit < 1.0.0 (dropping further to 3.13\% for Qiskit $\geq1.0.0$) despite performing similarly to the best model in~\cite{QiskitHumanEval}.

The results of our empirical comparison underscore the general performance of C4Q.
We believe that the architecture $-$ including its template-based code generation and deterministic logical engine $-$ is a key enable of both correctness and maintainability, particularly in handling newer Qiskit versions. 
The head-to-head comparison reinforces C4Q’s trustworthiness, and that its tailored architectural decisions sustain high-quality answers and streamline future adaptations.

%% file: 6-conclusions-and-future-work.tex
\section{Conclusions and Future Work} \label{s:conclusions}

The most frustrating part of working with LLMs is when they are wrong and we do not even realize it.
In the case of advanced topics that do not have much data available for training, even the most recent LLMs do not to produce answers with a quality that is good enough.
Instead of relying on an all-AI approach, C4Q 2.0's architecture uses LLMs for natural language processing, which they are treating very well, as well as specialized modules for generating answers. 
This allows C4Q 2.0 to answer questions accurately, even if limited to its current modules.
The result is a chatbot whose answers you can trust, enabling its use for teaching and learning in general.


In C4Q 2.0, the core features of C4Q 1.0 $-$ defining a quantum gate, providing its circuit representation, and applying it to a quantum state $-$ now produce not only the calculated results but also corresponding ready-to-run Qiskit code, which enables users to reproduce and further explore these outcomes practically. 
C4Q 2.0 can solve the TSP and the KP, with generated ready-to-run Qiskit code available for both, bridging the gap between introductory quantum topics and real-world optimization challenges.
Although parameter extraction for phase shifts and KP values exhibits higher error rates, C4Q architecture provides safety mechanisms (e.g., user confirmations) to maintain trustworthy outputs.
An empirical comparison with three existing chatbots further revealed that C4Q~2.0’s template-based approach and deterministic logical engine sustain higher accuracy, particularly when working with newer versions of Qiskit.


In the future, we aim to focus on improving the QA LLM's performance, particularly in accurately extracting phase shift parameters and interpreting KP-related parameters. 
We plan to collaborate with quantum computing educators to conduct user studies, ensuring that C4Q~2.0 meets pedagogical needs and supports quantum education more effectively. 
Additionally, implementing session-based accounts will allow users to maintain conversation histories, fostering better usability, traceability, and auditability.
These future developments will reinforce C4Q~2.0’s position as a reliable quantum software tool, aligning with broader requirements for transparent and evolving software architectures. 